\documentclass[sigconf]{acmart}

\usepackage[most]{tcolorbox}

\usepackage[normalem]{ulem}
\usepackage{tabularray}
\usepackage{makecell} 
\usepackage[normalem]{ulem}
\useunder{\uline}{\ul}{}
\usepackage{algorithm}
\usepackage{algorithmic}
\usepackage{multirow}
\usepackage{amsmath} 
\usepackage{enumitem}
\usepackage{balance}
\usepackage{caption} % 载入caption宏包

\usepackage{array} 
\usepackage{etoolbox} % 载入etoolbox宏包
\usepackage{graphicx}
\usepackage{booktabs}

\copyrightyear{2025}
\acmYear{2025}
\setcopyright{cc}
\setcctype{by-sa}
\acmConference[SIGIR '25]{Proceedings of the 48th International ACM SIGIR Conference on Research and Development in Information Retrieval}{July 13--18, 2025}{Padua, Italy}
\acmBooktitle{Proceedings of the 48th International ACM SIGIR Conference on Research and Development in Information Retrieval (SIGIR '25), July 13--18, 2025, Padua, Italy}\acmDOI{10.1145/3726302.3731692}
\acmISBN{979-8-4007-1592-1/2025/07}

\begin{document}
\title{Dynamic and Parametric Retrieval-Augmented Generation} 
\author{Weihang Su}
\email{swh22@mails.tsinghua.edu.cn}
% \affiliation{Department of Computer Science and Technology, Tsinghua University}
\affiliation{
DCST, Tsinghua University\\
Beijing 100084 \country{China}
}

\author{Qingyao Ai}
% \authornote{Corresponding author}
\email{aiqy@tsinghua.edu.cn}
\affiliation{
DCST, Tsinghua University\\
Beijing 100084 \country{China}
}

\author{Jingtao Zhan}
\email{jingtaozhan@gmail.com}
\affiliation{
DCST, Tsinghua University\\
Beijing 100084 \country{China}
}

\author{Qian Dong}
\email{dq22@mails.tsinghua.edu.cn}
\affiliation{
DCST, Tsinghua University\\
Beijing 100084 \country{China}
}

\author{Yiqun Liu}
\email{yiqunliu@tsinghua.edu.cn	}
\affiliation{
DCST, Tsinghua University\\
Beijing 100084 \country{China}
}

\renewcommand{\shortauthors}{Weihang Su, Qingyao Ai, Jingtao Zhan, Qian Dong, and Yiqun Liu}
\begin{abstract}

Retrieval-Augmented Generation (RAG) has become a foundational paradigm for equipping large language models (LLMs) with external knowledge, playing a critical role in information retrieval and knowledge-intensive applications. 
However, conventional RAG systems typically adopt a static retrieve-then-generate pipeline and rely on in-context knowledge injection, which can be suboptimal for complex tasks that require multihop reasoning, adaptive information access, and deeper integration of external knowledge. 
Motivated by these limitations, the research community has moved beyond static retrieval and in-context knowledge injection. 
Among the emerging directions, this tutorial delves into two rapidly growing and complementary research areas on RAG: Dynamic RAG and Parametric RAG.
Dynamic RAG adaptively determines when and what to retrieve during the LLM's generation process, enabling real-time adaptation to the LLM's evolving information needs. 
Parametric RAG rethinks how retrieved knowledge should be injected into LLMs, transitioning from input-level to parameter-level knowledge injection for enhanced efficiency and effectiveness. 
This tutorial offers a comprehensive overview of recent advances in these emerging research areas. It also shares theoretical foundations and practical insights to support and inspire further research in RAG.

\end{abstract}

\maketitle
\pagestyle{plain}

\section*{Cover Sheet Information}
% \vspace{2mm}
\noindent \textbf{Title:} Dynamic and Parametric Retrieval-Augmented Generation 
\vspace{1mm}

\noindent \textbf{Length:} Half-day (3 hours) 
\vspace{1mm}

\noindent \textbf{Format:} This is a half-day (3-hour) lecture-style tutorial, including scheduled breaks. It will be conducted on-site, with at least two presenters planning to attend the conference in person.
\vspace{1mm}

\noindent \textbf{Intended Audience:} Intermediate. This tutorial is intended for researchers, practitioners, and students who have prior experience with information retrieval or natural language processing and are interested in the intersection of IR techniques and large language models. It is particularly suitable for those seeking to understand recent advances in retrieval-augmented generation (RAG) and knowledge injection within large language models (LLMs). 
\vspace{1mm}

\noindent \textbf{Prerequisite Knowledge:} Attendees are expected to have a basic understanding of the Transformer-based large language models~\cite{brown2020language,fang2024scaling,touvron2023llama,yang2024qwen2,liu2024deepseek}, including concepts such as attention mechanisms. 
In addition, familiarity with the fundamentals of both sparse~\cite{zhai2008statistical,robertson2009probabilistic} and dense retrieval~\cite{karpukhin2020dense,zhan2021optimizing,su2024wikiformer,gao2021condenser} will be helpful. 
\vspace{1mm}

\noindent \textbf{Previous Talks:} This tutorial has not been presented before. 

\vspace{1mm}

\noindent
\textbf{Presenters:} 

\vspace{1mm}

\noindent \textbf{Weihang Su}\footnote{Main Contact Person: swh22@mails.tsinghua.edu.cn}~\cite{oneal2000homepage} is a Ph.D. student at the Department of Computer Science and Technology, Tsinghua University, supervised by Prof. Yiqun Liu. His research focuses on Retrieval-Augmented Generation (RAG) and knowledge injection for LLMs. He has published papers at top-tier conferences and journals, including AAAI, ACL, EMNLP, WebConf, SIGIR, and TOIS. He serves as a reviewer or PC member for major conferences such as SIGIR, WebConf, EMNLP, NeurIPS, ICLR, ICML, ACL, and CIKM.

\vspace{1mm}

% \noindent \textbf{Qingyao Ai}\footnote{Email: aiqy@tsinghua.edu.cn}~\cite{qingyaoaihomepage} is an Associate Professor at the Department of Computer Science and Technology, Tsinghua University. His research lies at the intersection of information retrieval and generative AI, with a focus on retrieval/ranking optimization, retrieval-augmented generation, and agent-oriented prompt learning. He received his Ph.D. from the University of Massachusetts Amherst under the supervision of Prof. W. Bruce Croft. He has served in various academic roles, including General Co-Chair of SIGIR-AP 2023 and Program Co-Chair of NTCIR-18, and is currently an Associate Editor of ACM TOIS. He has also been an Area Chair or Senior PC Member for SIGIR, WWW, CIKM, NAACL, and EMNLP.

\noindent \textbf{Qingyao Ai}\footnote{Email: aiqy@tsinghua.edu.cn}~\cite{qingyaoaihomepage} is an Associate Professor at the Department of Computer Science and Technology, Tsinghua University. His research lies at the intersection of information retrieval and generative AI, with a focus on retrieval/ranking optimization and retrieval-augmented generation. 
He received his Ph.D. from the University of Massachusetts Amherst under the supervision of Prof. W. Bruce Croft.
He has served in various academic roles, including General Co-Chair of SIGIR-AP 2023, Program Co-Chair of NTCIR-18, and is currently an Associate Editor of ACM TOIS. 

\vspace{1mm}

% \noindent \textbf{Jingtao Zhan}\footnote{Email: jingtaozhan@gmail.com}~\cite{jingtaozhanhomepage} is a Ph.D. student at the Department of Computer Science and Technology, Tsinghua University, advised by Prof. Shaoping Ma and Prof. Yiqun Liu. His research centers on building intelligent information systems, with a focus on both efficient information retrieval and improving AIGC systems through user interaction logs. 
% His work spans evaluation scheme design, training frameworks, and index optimization, with one paper winning the Best Paper Award at WSDM 2022. More recently, his work on leveraging interaction logs for prompt optimization in AIGC systems has been accepted by SIGIR 2024. He has served as a reviewer or PC member for major conferences including SIGIR, ACL, TheWebConf, WSDM, CIKM, and COLING.
\noindent \textbf{Jingtao Zhan}\footnote{Email: jingtaozhan@gmail.com}~\cite{jingtaozhanhomepage} is a Ph.D. student at Tsinghua University, advised by Prof. Shaoping Ma and Prof. Yiqun Liu. His research focuses on efficient information retrieval and improving AIGC systems using user interaction logs. 
He has published papers at top-tier conferences and received multiple honors, including the Best Paper Awards at WSDM 2022 and SIGIR 2024. He also serves as a reviewer for leading venues such as SIGIR, ACL, and The Web Conference.

\section{Motivation and Overview}

Large Language Models (LLMs) have demonstrated remarkable progress across a wide range of natural language processing (NLP) and information retrieval (IR) tasks~\cite{brown2020language,fang2024scaling,touvron2023llama,yang2024qwen2,liu2024deepseek}. 
Despite these advancements, they face challenges such as hallucinations~\cite{ji2023survey,su2024unsupervised,rawte2023survey,wang2025joint}, outdated knowledge~\cite{wang2024knowledgesurvey,wang2025decouple,wang2024lekube,fang2024alphaedit,wang2024knowledge}, and domain adaptation~\cite{gururangan2020don,su2025judge,su2023caseformer,su2023legalaid,su2024stard}.
To address these challenges, Retrieval-Augmented Generation (RAG)~\cite{borgeaud2022improving,lewis2020retrieval,su2024mitigating,guu2020retrieval,izacard2020leveraging,jiang2022retrieval,tu2025rbft,dong2025decoupling} has emerged as a promising solution that supplements LLMs with access to external knowledge.
The conventional RAG paradigm adopts a straightforward retrieve-then-generate approach: an external retriever~\cite{zhai2008statistical,gao2021condenser,su2023thuir2,ma2023caseencoder} or a more complex retrieval system~\cite{salemi2024towards,li2023towards,ye2024relevance,chen2022web} retrieves relevant documents based on the user's query, then these documents are appended into the contextual input of the LLM, which subsequently generates a response grounded in the retrieved evidence. 
This paradigm has proven effective across a variety of knowledge-intensive applications and has been widely adopted in practice.

\subsection{Limitations of Standard RAG}
\label{sec-limitation}

In standard RAG, retrieval is performed before the generation process, based on the initial user query. 
This one-time retrieval strategy assumes that all relevant information can be collected in advance. 
However, for tasks that involve multi-hop reasoning or long-form generation, the model’s information needs may evolve during the generation process. 
For example, intermediate reasoning steps may introduce new sub-questions or require clarification of previously unseen concepts. 
To support such scenarios, retrieval must be dynamically integrated with generation, allowing the model to access information during the generation process. Since standard RAG performs retrieval before generation, it often fails to provide all the knowledge required by the LLM during the generation process.

Another fundamental limitation of standard RAG lies in how the retrieved knowledge is integrated into the model.
Specifically, standard RAG adopts the \textit{in-context knowledge injection} approach, where the retrieved documents are appended to the input prompt and jointly processed with the query.
While this approach is straightforward to implement, it introduces several critical challenges.
First, when the amount of relevant knowledge is substantial, appending it directly to the input significantly increases the context length, leading to higher computational costs and latency during inference.
Additionally, as the context becomes longer, the model's attention becomes more dispersed, reducing its ability to focus on the most relevant information~\cite{liu2024lost,levy2024same}.
More fundamentally, the way LLMs utilize in-context information is inherently different from how they leverage internal parametric knowledge. 
Empirical studies show that LLMs encode most of their factual knowledge within the parameters of their neural architecture, particularly in the feed-forward network layers~\cite{yu-ananiadou-2024-neuron,nanda2023fact,fang2024alphaedit}. 
In contrast, information provided in the input context only influences the model through the dynamic computation of key-value pairs in the self-attention mechanism, without being integrated into the model’s internal parametric knowledge representations. 
This fundamental difference in how internal and contextual knowledge are processed leads to a weaker and less reliable use of external knowledge: LLMs may fail to consistently ground their outputs in the retrieved content, especially when the knowledge conflicts with their internal parametric knowledge.

These limitations have motivated the development of more advanced RAG paradigms that aim to overcome the constraints of static retrieval and the limitations of in-context knowledge injection. 
Two representative directions have recently gained significant attention: \emph{Dynamic RAG} and \emph{Parametric RAG}.
Dynamic RAG enables retrieval to be dynamically integrated with generation, allowing the system to access relevant information on demand based on the evolving context, which is particularly beneficial for multi-hop reasoning and complex generation tasks.
From another angle, Parametric RAG seeks to inject external knowledge directly into the model’s parameters, enabling deeper integration and allowing the LLM to utilize external knowledge in the same way it leverages its internal parametric knowledge.

This tutorial provides a comprehensive introduction to these emerging RAG paradigms, equipping participants with theoretical insights and practical techniques for building next-generation retrieval-augmented systems.

% In summary, the limitations of standard RAG arise from two primary factors: its static retrieval strategy and its shallow integration of external knowledge.
% Overcoming these limitations calls for retrieval mechanisms that dynamically adapt to the model's evolving information needs, as well as more effective knowledge injection methods that directly embed external information into the LLM's internal parametric knowledge.
% These limitations have motivated two emerging research directions: Dynamic RAG, which supports active retrieval and generation, and Parametric RAG, which injects the retrieved knowledge directly into the model’s parameters.

\subsection{Dynamic Retrieval-Augmented Generation}

Dynamic RAG is an emerging paradigm that overcomes the limitations of standard RAG. 
Unlike standard RAG, which relies on a single retrieval step triggered by the initial query, Dynamic RAG supports multiple rounds of retrieval that adapt to the evolving information needs of an LLM during its generation process~\cite{asai2024selfrag,su2024dragin,jiang2023active,yao2024seakr,liu2024ctrla}. At each retrieval step, the decision of \textit{when} and \textit{what} to retrieve can either be made by the LLM itself (e.g., by generating special tokens indicating the need for external knowledge~\cite{asai2024selfrag}) or by an external system that monitors the model's generation state to detect uncertainty~\cite{su2024dragin,jiang2023active}. 
Similarly, retrieval queries can be formulated either directly by the LLM or through an external query-generation module. 
By iteratively incorporating retrieved external knowledge at relevant points throughout generation, Dynamic RAG enables the model to generate more accurate, context-aware, and comprehensive responses, significantly enhancing performance on complex tasks such as multi-hop reasoning and long-form text generation.

Recent advancements have demonstrated the effectiveness of this paradigm. 
% For instance, \textbf{Self-Reflective RAG} (Self-RAG)~\cite{asai2024selfrag} introduces “reflection tokens” as signals of uncertainty within the model’s output stream. 
For instance, \textbf{Self-Reflective RAG} (Self-RAG)~\cite{asai2024selfrag} introduces “reflection tokens” as explicit signals indicating that the model requires external knowledge to assist generation.
These tokens trigger the system to perform on-the-fly retrieval, so that the LLM can actively retrieve external information as required.
Experimental results show that this strategy outperforms static RAG, especially on tasks that require precise factual grounding.
\textbf{FLARE}~\cite{jiang2023active} introduces an uncertainty-aware retrieval mechanism that monitors token-level generation probabilities during decoding. When low-confidence tokens are detected, the system automatically formulates retrieval queries and updates the generation context with the retrieved information. This mechanism enables the model to correct potential hallucinations and improve factual accuracy.
\textbf{DRAGIN}~\cite{su2024dragin} improves upon FLARE by proposing a lightweight RAG framework that dynamically determines when and what to retrieve based on the LLM’s information needs during generation. Unlike Self-RAG and FLARE, DRAGIN requires no additional training or prompt modifications. It models retrieval needs by analyzing the model’s internal signals, such as attention distributions and predictive entropy, and formulates precise retrieval queries using self-attention over the current context. Experimental results show that DRAGIN achieves superior performance on complex tasks that demand dynamic access to external knowledge.
\textbf{SeaKR}~\cite{yao2024seakr} introduces a self-aware retrieval mechanism that leverages the LLM's internal states to quantify token-level uncertainty. When the uncertainty surpasses a predefined threshold, the system triggers external retrieval and re-ranks the retrieved passages based on their potential to reduce uncertainty. This targeted retrieval approach improves factual accuracy while avoiding unnecessary overhead.

In summary, Dynamic RAG offers a more flexible alternative to standard RAG by enabling the model to retrieve information in an on-demand, iterative manner. 
Through mechanisms such as self-reflection, uncertainty detection, and token-level influence analysis, Dynamic RAG allows the model to adaptively identify when retrieval is necessary and what information to retrieve when the retrieval module is triggered. 
This dynamic interaction between generation and retrieval not only improves factual consistency but also enhances the model's ability to handle complex, multi-step reasoning tasks that cannot be effectively supported by static, one-shot retrieval.

\subsection{Parametric Retrieval-Augmented Generation}

While Dynamic RAG addresses the question of {when} and {what} to retrieve during generation, Parametric Retrieval-Augmented Generation (Parametric RAG) focuses on a more fundamental challenge: \emph{how to integrate retrieved knowledge into the model}. 
Despite the growing body of work in the RAG literature, the vast majority of existing approaches adopt in-context knowledge injection, where external passages are appended to the input prompt and processed jointly with the query.
However, as discussed in \S~\ref{sec-limitation}, this approach suffers from notable limitations in both efficiency and effectiveness~\cite{liu2024lost,levy2024same,yu-ananiadou-2024-neuron}.
To overcome these limitations, Parametric RAG introduces an alternative paradigm that integrates retrieved knowledge directly into the model’s parameters.
The core challenge of Parametric RAG is to devise a transformation that maps a document $D$ into a plug-in parameter module $P$, such that, once $P$ is inserted into the LLM, the model can internalize the knowledge embedded in $D$.
To address this challenge, two mainstream approaches have been proposed for implementing the Document-to-Parameter transformation.

The first class of methods follows a training-based approach to obtain document-specific parametric representations. 
For each document, a synthetic dataset is constructed by expanding the original content, typically through rewriting or generating QA pairs. 
Lightweight adapter modules, such as LoRA~\cite{hulora}, are then fine-tuned on this dataset. 
The resulting LoRA modules function as plug-in parameters that encode the document’s knowledge and serve as its parametric representation.
A representative method in this category is \textbf{PRAG}~\cite{su2025paramrag}, which adopts a two-stage pipeline: during the offline phase, each document is encoded into a document-specific LoRA module via the aforementioned training process; at inference time, the top-$k$ relevant modules are retrieved and merged into the base model, enabling the updated LLM to perform query-specific generation.

The second class of methods eliminates the need for document-specific fine-tuning or storage by leveraging an online parameter generation strategy. Instead of pre-training individual adapters, a small hypernetwork is trained to map each document's semantic embedding into a parameter-efficient module. At inference time, the retrieved document's semantic embedding is passed through this hypernetwork to dynamically generate plug-in parameters on-the-fly. 
This approach enables real-time adaptation and significantly reduces the cost of maintaining a large corpus of parametric modules. 
\textbf{DyPRAG}~\cite{tan2025better} exemplifies this approach by introducing a dynamic parameter translator that generates document-specific parametric representations on-the-fly, conditioned on the retrieved document’s semantic embedding. This design enables Parametric RAG without maintaining a large corpus of parametric modules, thereby achieving both scalability and efficiency.

In summary, Parametric RAG represents a paradigm shift from context to parameter-level knowledge injection, providing a more effective solution for integrating external information into LLMs.

\section{Objectives}

% This tutorial aims to equip attendees with essential concepts, practical techniques, and insights to effectively leverage Dynamic and Parametric Retrieval-Augmented Generation. After attending this tutorial, participants will be able to:

This tutorial aims to provide attendees with a comprehensive understanding of recent advancements in Retrieval-Augmented Generation (RAG), with a particular focus on two emerging and complementary paradigms: Dynamic RAG and Parametric RAG. By attending this tutorial, participants will:

\begin{itemize}[leftmargin=*]
\item  Gain a clear understanding of the fundamentals of RAG, including its current landscape and the key limitations of standard RAG frameworks.

\item Understand the motivation, architecture, and representative techniques of Dynamic RAG, which interleaves retrieval with generation to support real-time, adaptive access to external knowledge.

\item Learn the principles and implementation strategies of Parametric RAG, an emerging RAG paradigm that injects external knowledge directly into model parameters.

\item Gain insights into emerging research trends, open challenges, and future directions in the evolving RAG landscape.

\end{itemize}

\section{Relevance to IR Community}

Retrieval-Augmented Generation (RAG) directly addresses critical challenges in modern information retrieval by combining traditional retrieval techniques with powerful language generation capabilities of LLMs.
Unlike classical IR methods that primarily focus on retrieving static documents in response to a single query, many emerging RAG frameworks in recent years introduce dynamic interactions between retrieval and generation, enabling systems to more effectively address complex queries and support multihop reasoning.
This tutorial targets core challenges at the intersection of retrieval and generation, with a particular emphasis on two emerging and complementary research directions: \emph{Dynamic RAG} and \emph{Parametric RAG}.
Dynamic RAG investigates when and what to retrieve by interleaving retrieval with the generation process, thereby addressing the limitations of one-shot retrieval strategies.
Parametric RAG explores how retrieved knowledge can be encoded into the model’s parameters, enabling more effective knowledge injection.
Both directions represent critical advancements for the IR community, as they offer novel solutions for building more adaptive, accurate, and efficient retrieval-augmented generation systems.

While prior tutorials at SIGIR 2022~\cite{cai2022recent} and ACL 2023~\cite{asai2023retrieval} have introduced advances on standard RAG methods, they do not discuss the emerging RAG paradigms based on dynamic retrieval or parametric knowledge injection. 
This tutorial fills that gap by providing practical insights and technical depth on these two complementary RAG paradigms, offering insights for IR researchers to develop next-generation IR systems.

\section{Format and Schedule}

This is a half-day (3-hour) lecture-style tutorial, including scheduled breaks. It will be conducted on-site, with at least two presenters planning to attend the conference in person.

\begin{itemize}[leftmargin=*]
    \item \textbf{Introduction and Background} (30 min): 

     \begin{itemize}
        \item Tutorial Overview
        \item Foundations of RAG
        \item New RAG Paradigms
        \item Limitations of Standard RAG
    \end{itemize}

\vspace{1mm}

    \item \textbf{Topic 1 — Dynamic RAG} (50 min): 
    \begin{itemize}
\item Motivation: Why is Dynamic RAG necessary?

\item When to Retrieve: When should the retrieval module be triggered during the generation process?

\item What to Retrieve: How can we formulate queries that reflect the model’s real-time information needs?

\item Open Challenges and Future Directions
    \end{itemize}

\vspace{1mm}

    \item \textbf{Q\&A} (10 min)
\vspace{1mm}

    \item \textbf{Break} (30 min)

\vspace{1mm}

    \item \textbf{Topic 2 — Parametric RAG} (40 min):
    
\begin{itemize}

\item Background: Overview of existing works on how LLMs encode world knowledge in their parameters.

\item Motivation: Discussion of scenarios where in-context knowledge injection faces limitations.

\item Parametric Knowledge Representation: Summary of methods for converting textual knowledge into plug-in parameters.

\item Existing Works on Parametric RAG: Introduction of representative Parametric RAG approaches and their design choices.
        
    \item Open Challenges and Future Directions
        
\end{itemize}

\vspace{1mm}

    \item \textbf{Summary and Future Directions} (10 min): A summary of the tutorial’s key insights, discussion of open problems, and future research opportunities in retrieval-augmented generation.
\vspace{1mm}

    \item \textbf{Q\&A} (10 min)

\end{itemize}

\section{Supporting Materials}

The supplemental materials of this tutorial will include a manuscript that describes the details of the technical content, as well as a comprehensive list of important references and related tutorials grouped by topic. The tutorial slides will also be provided.
In addition, we will maintain a GitHub repository that surveys and categorizes all relevant papers related to this topic to facilitate further reading and exploration. All materials will be made accessible through a dedicated website, which will be shared with the attendees before the tutorial.

\bibliographystyle{ACM-Reference-Format}
% \balance
\bibliography{sample-base}


\begin{thebibliography}{56}

%%% the Web address.

\ifx \showCODEN    \undefined \def \showCODEN     #1{\unskip}     \fi
\ifx \showDOI      \undefined \def \showDOI       #1{#1}\fi
\ifx \showISBNx    \undefined \def \showISBNx     #1{\unskip}     \fi
\ifx \showISBNxiii \undefined \def \showISBNxiii  #1{\unskip}     \fi
\ifx \showISSN     \undefined \def \showISSN      #1{\unskip}     \fi
\ifx \showLCCN     \undefined \def \showLCCN      #1{\unskip}     \fi
\ifx \shownote     \undefined \def \shownote      #1{#1}          \fi
\ifx \showarticletitle \undefined \def \showarticletitle #1{#1}   \fi
\ifx \showURL      \undefined \def \showURL       {\relax}        \fi
% invisible to TeX
\providecommand\bibfield[2]{#2}
\providecommand\bibinfo[2]{#2}
\providecommand\natexlab[1]{#1}
\providecommand\showeprint[2][]{arXiv:#2}

\bibitem[Ai(2025)]%
        {qingyaoaihomepage}
\bibfield{author}{\bibinfo{person}{Qingyao Ai}.} \bibinfo{year}{2025}\natexlab{}.
\newblock \bibinfo{title}{Homepage of Qingyao Ai}.
\newblock \bibinfo{howpublished}{\url{https://qingyaoai.github.io/}}.
\newblock
\newblock
\shownote{Accessed: 2025-03-28}.


\bibitem[Asai et~al\mbox{.}(2023)]%
        {asai2023retrieval}
\bibfield{author}{\bibinfo{person}{Akari Asai}, \bibinfo{person}{Sewon Min}, \bibinfo{person}{Zexuan Zhong}, {and} \bibinfo{person}{Danqi Chen}.} \bibinfo{year}{2023}\natexlab{}.
\newblock \showarticletitle{Retrieval-based language models and applications}. In \bibinfo{booktitle}{\emph{Proceedings of the 61st Annual Meeting of the Association for Computational Linguistics (Volume 6: Tutorial Abstracts)}}. \bibinfo{pages}{41--46}.
\newblock


\bibitem[Asai et~al\mbox{.}(2024)]%
        {asai2024selfrag}
\bibfield{author}{\bibinfo{person}{Akari Asai}, \bibinfo{person}{Zeqiu Wu}, \bibinfo{person}{Yizhong Wang}, \bibinfo{person}{Avirup Sil}, {and} \bibinfo{person}{Hannaneh Hajishirzi}.} \bibinfo{year}{2024}\natexlab{}.
\newblock \showarticletitle{Self-Reflective Retrieval-Augmented Generation (Self-RAG)}. In \bibinfo{booktitle}{\emph{International Conference on Learning Representations (ICLR)}}.
\newblock


\bibitem[Borgeaud et~al\mbox{.}(2022)]%
        {borgeaud2022improving}
\bibfield{author}{\bibinfo{person}{Sebastian Borgeaud}, \bibinfo{person}{Arthur Mensch}, \bibinfo{person}{Jordan Hoffmann}, \bibinfo{person}{Trevor Cai}, \bibinfo{person}{Eliza Rutherford}, \bibinfo{person}{Katie Millican}, \bibinfo{person}{George~Bm Van Den~Driessche}, \bibinfo{person}{Jean-Baptiste Lespiau}, \bibinfo{person}{Bogdan Damoc}, \bibinfo{person}{Aidan Clark}, {et~al\mbox{.}}} \bibinfo{year}{2022}\natexlab{}.
\newblock \showarticletitle{Improving language models by retrieving from trillions of tokens}. In \bibinfo{booktitle}{\emph{International conference on machine learning}}. PMLR, \bibinfo{pages}{2206--2240}.
\newblock


\bibitem[Brown et~al\mbox{.}(2020)]%
        {brown2020language}
\bibfield{author}{\bibinfo{person}{Tom Brown}, \bibinfo{person}{Benjamin Mann}, \bibinfo{person}{Nick Ryder}, \bibinfo{person}{Melanie Subbiah}, \bibinfo{person}{Jared~D Kaplan}, \bibinfo{person}{Prafulla Dhariwal}, \bibinfo{person}{Arvind Neelakantan}, \bibinfo{person}{Pranav Shyam}, \bibinfo{person}{Girish Sastry}, \bibinfo{person}{Amanda Askell}, {et~al\mbox{.}}} \bibinfo{year}{2020}\natexlab{}.
\newblock \showarticletitle{Language models are few-shot learners}.
\newblock \bibinfo{journal}{\emph{Advances in neural information processing systems}}  \bibinfo{volume}{33} (\bibinfo{year}{2020}), \bibinfo{pages}{1877--1901}.
\newblock


\bibitem[Cai et~al\mbox{.}(2022)]%
        {cai2022recent}
\bibfield{author}{\bibinfo{person}{Deng Cai}, \bibinfo{person}{Yan Wang}, \bibinfo{person}{Lemao Liu}, {and} \bibinfo{person}{Shuming Shi}.} \bibinfo{year}{2022}\natexlab{}.
\newblock \showarticletitle{Recent advances in retrieval-augmented text generation}. In \bibinfo{booktitle}{\emph{Proceedings of the 45th international ACM SIGIR conference on research and development in information retrieval}}. \bibinfo{pages}{3417--3419}.
\newblock


\bibitem[Chen et~al\mbox{.}(2022)]%
        {chen2022web}
\bibfield{author}{\bibinfo{person}{Xuesong Chen}, \bibinfo{person}{Ziyi Ye}, \bibinfo{person}{Xiaohui Xie}, \bibinfo{person}{Yiqun Liu}, \bibinfo{person}{Xiaorong Gao}, \bibinfo{person}{Weihang Su}, \bibinfo{person}{Shuqi Zhu}, \bibinfo{person}{Yike Sun}, \bibinfo{person}{Min Zhang}, {and} \bibinfo{person}{Shaoping Ma}.} \bibinfo{year}{2022}\natexlab{}.
\newblock \showarticletitle{Web search via an efficient and effective brain-machine interface}. In \bibinfo{booktitle}{\emph{Proceedings of the fifteenth ACM international conference on web search and data mining}}. \bibinfo{pages}{1569--1572}.
\newblock


\bibitem[Dong et~al\mbox{.}(2025)]%
        {dong2025decoupling}
\bibfield{author}{\bibinfo{person}{Qian Dong}, \bibinfo{person}{Qingyao Ai}, \bibinfo{person}{Hongning Wang}, \bibinfo{person}{Yiding Liu}, \bibinfo{person}{Haitao Li}, \bibinfo{person}{Weihang Su}, \bibinfo{person}{Yiqun Liu}, \bibinfo{person}{Tat-Seng Chua}, {and} \bibinfo{person}{Shaoping Ma}.} \bibinfo{year}{2025}\natexlab{}.
\newblock \showarticletitle{Decoupling Knowledge and Context: An Efficient and Effective Retrieval Augmented Generation Framework via Cross Attention}. In \bibinfo{booktitle}{\emph{Proceedings of the ACM on Web Conference 2025}}. \bibinfo{pages}{4386--4395}.
\newblock


\bibitem[Fang et~al\mbox{.}(2024a)]%
        {fang2024alphaedit}
\bibfield{author}{\bibinfo{person}{Junfeng Fang}, \bibinfo{person}{Houcheng Jiang}, \bibinfo{person}{Kun Wang}, \bibinfo{person}{Yunshan Ma}, \bibinfo{person}{Shi Jie}, \bibinfo{person}{Xiang Wang}, \bibinfo{person}{Xiangnan He}, {and} \bibinfo{person}{Tat-Seng Chua}.} \bibinfo{year}{2024}\natexlab{a}.
\newblock \showarticletitle{Alphaedit: Null-space constrained knowledge editing for language models}.
\newblock \bibinfo{journal}{\emph{arXiv preprint arXiv:2410.02355}} (\bibinfo{year}{2024}).
\newblock


\bibitem[Fang et~al\mbox{.}(2024b)]%
        {fang2024scaling}
\bibfield{author}{\bibinfo{person}{Yan Fang}, \bibinfo{person}{Jingtao Zhan}, \bibinfo{person}{Qingyao Ai}, \bibinfo{person}{Jiaxin Mao}, \bibinfo{person}{Weihang Su}, \bibinfo{person}{Jia Chen}, {and} \bibinfo{person}{Yiqun Liu}.} \bibinfo{year}{2024}\natexlab{b}.
\newblock \showarticletitle{Scaling laws for dense retrieval}. In \bibinfo{booktitle}{\emph{Proceedings of the 47th International ACM SIGIR Conference on Research and Development in Information Retrieval}}. \bibinfo{pages}{1339--1349}.
\newblock


\bibitem[Gao and Callan(2021)]%
        {gao2021condenser}
\bibfield{author}{\bibinfo{person}{Luyu Gao} {and} \bibinfo{person}{Jamie Callan}.} \bibinfo{year}{2021}\natexlab{}.
\newblock \showarticletitle{Condenser: a pre-training architecture for dense retrieval}.
\newblock \bibinfo{journal}{\emph{arXiv preprint arXiv:2104.08253}} (\bibinfo{year}{2021}).
\newblock


\bibitem[Gururangan et~al\mbox{.}(2020)]%
        {gururangan2020don}
\bibfield{author}{\bibinfo{person}{Suchin Gururangan}, \bibinfo{person}{Ana Marasovi{\'c}}, \bibinfo{person}{Swabha Swayamdipta}, \bibinfo{person}{Kyle Lo}, \bibinfo{person}{Iz Beltagy}, \bibinfo{person}{Doug Downey}, {and} \bibinfo{person}{Noah~A Smith}.} \bibinfo{year}{2020}\natexlab{}.
\newblock \showarticletitle{Don't stop pretraining: Adapt language models to domains and tasks}.
\newblock \bibinfo{journal}{\emph{arXiv preprint arXiv:2004.10964}} (\bibinfo{year}{2020}).
\newblock


\bibitem[Guu et~al\mbox{.}(2020)]%
        {guu2020retrieval}
\bibfield{author}{\bibinfo{person}{Kelvin Guu}, \bibinfo{person}{Kenton Lee}, \bibinfo{person}{Zora Tung}, \bibinfo{person}{Panupong Pasupat}, {and} \bibinfo{person}{Mingwei Chang}.} \bibinfo{year}{2020}\natexlab{}.
\newblock \showarticletitle{Retrieval augmented language model pre-training}. In \bibinfo{booktitle}{\emph{International conference on machine learning}}. PMLR, \bibinfo{pages}{3929--3938}.
\newblock


\bibitem[Hu et~al\mbox{.}(2022)]%
        {hulora}
\bibfield{author}{\bibinfo{person}{Edward~J Hu}, \bibinfo{person}{Phillip Wallis}, \bibinfo{person}{Zeyuan Allen-Zhu}, \bibinfo{person}{Yuanzhi Li}, \bibinfo{person}{Shean Wang}, \bibinfo{person}{Lu Wang}, \bibinfo{person}{Weizhu Chen}, {et~al\mbox{.}}} \bibinfo{year}{2022}\natexlab{}.
\newblock \showarticletitle{LoRA: Low-Rank Adaptation of Large Language Models}. In \bibinfo{booktitle}{\emph{International Conference on Learning Representations}}.
\newblock


\bibitem[Izacard and Grave(2020)]%
        {izacard2020leveraging}
\bibfield{author}{\bibinfo{person}{Gautier Izacard} {and} \bibinfo{person}{Edouard Grave}.} \bibinfo{year}{2020}\natexlab{}.
\newblock \showarticletitle{Leveraging passage retrieval with generative models for open domain question answering}.
\newblock \bibinfo{journal}{\emph{arXiv preprint arXiv:2007.01282}} (\bibinfo{year}{2020}).
\newblock


\bibitem[Ji et~al\mbox{.}(2023)]%
        {ji2023survey}
\bibfield{author}{\bibinfo{person}{Ziwei Ji}, \bibinfo{person}{Nayeon Lee}, \bibinfo{person}{Rita Frieske}, \bibinfo{person}{Tiezheng Yu}, \bibinfo{person}{Dan Su}, \bibinfo{person}{Yan Xu}, \bibinfo{person}{Etsuko Ishii}, \bibinfo{person}{Ye~Jin Bang}, \bibinfo{person}{Andrea Madotto}, {and} \bibinfo{person}{Pascale Fung}.} \bibinfo{year}{2023}\natexlab{}.
\newblock \showarticletitle{Survey of hallucination in natural language generation}.
\newblock \bibinfo{journal}{\emph{Comput. Surveys}} \bibinfo{volume}{55}, \bibinfo{number}{12} (\bibinfo{year}{2023}), \bibinfo{pages}{1--38}.
\newblock


\bibitem[Jiang et~al\mbox{.}(2022)]%
        {jiang2022retrieval}
\bibfield{author}{\bibinfo{person}{Zhengbao Jiang}, \bibinfo{person}{Luyu Gao}, \bibinfo{person}{Jun Araki}, \bibinfo{person}{Haibo Ding}, \bibinfo{person}{Zhiruo Wang}, \bibinfo{person}{Jamie Callan}, {and} \bibinfo{person}{Graham Neubig}.} \bibinfo{year}{2022}\natexlab{}.
\newblock \showarticletitle{Retrieval as attention: End-to-end learning of retrieval and reading within a single transformer}.
\newblock \bibinfo{journal}{\emph{arXiv preprint arXiv:2212.02027}} (\bibinfo{year}{2022}).
\newblock


\bibitem[Jiang et~al\mbox{.}(2023)]%
        {jiang2023active}
\bibfield{author}{\bibinfo{person}{Zhengbao Jiang}, \bibinfo{person}{Frank~F Xu}, \bibinfo{person}{Luyu Gao}, \bibinfo{person}{Zhiqing Sun}, \bibinfo{person}{Qian Liu}, \bibinfo{person}{Jane Dwivedi-Yu}, \bibinfo{person}{Yiming Yang}, \bibinfo{person}{Jamie Callan}, {and} \bibinfo{person}{Graham Neubig}.} \bibinfo{year}{2023}\natexlab{}.
\newblock \showarticletitle{Active retrieval augmented generation}.
\newblock \bibinfo{journal}{\emph{arXiv preprint arXiv:2305.06983}} (\bibinfo{year}{2023}).
\newblock


\bibitem[Karpukhin et~al\mbox{.}(2020)]%
        {karpukhin2020dense}
\bibfield{author}{\bibinfo{person}{Vladimir Karpukhin}, \bibinfo{person}{Barlas O{\u{g}}uz}, \bibinfo{person}{Sewon Min}, \bibinfo{person}{Patrick Lewis}, \bibinfo{person}{Ledell Wu}, \bibinfo{person}{Sergey Edunov}, \bibinfo{person}{Danqi Chen}, {and} \bibinfo{person}{Wen-tau Yih}.} \bibinfo{year}{2020}\natexlab{}.
\newblock \showarticletitle{Dense passage retrieval for open-domain question answering}.
\newblock \bibinfo{journal}{\emph{arXiv preprint arXiv:2004.04906}} (\bibinfo{year}{2020}).
\newblock


\bibitem[Levy et~al\mbox{.}(2024)]%
        {levy2024same}
\bibfield{author}{\bibinfo{person}{Mosh Levy}, \bibinfo{person}{Alon Jacoby}, {and} \bibinfo{person}{Yoav Goldberg}.} \bibinfo{year}{2024}\natexlab{}.
\newblock \showarticletitle{Same task, more tokens: the impact of input length on the reasoning performance of large language models}.
\newblock \bibinfo{journal}{\emph{arXiv preprint arXiv:2402.14848}} (\bibinfo{year}{2024}).
\newblock


\bibitem[Lewis et~al\mbox{.}(2020)]%
        {lewis2020retrieval}
\bibfield{author}{\bibinfo{person}{Patrick Lewis}, \bibinfo{person}{Ethan Perez}, \bibinfo{person}{Aleksandra Piktus}, \bibinfo{person}{Fabio Petroni}, \bibinfo{person}{Vladimir Karpukhin}, \bibinfo{person}{Naman Goyal}, \bibinfo{person}{Heinrich K{\"u}ttler}, \bibinfo{person}{Mike Lewis}, \bibinfo{person}{Wen-tau Yih}, \bibinfo{person}{Tim Rockt{\"a}schel}, {et~al\mbox{.}}} \bibinfo{year}{2020}\natexlab{}.
\newblock \showarticletitle{Retrieval-augmented generation for knowledge-intensive nlp tasks}.
\newblock \bibinfo{journal}{\emph{Advances in Neural Information Processing Systems}}  \bibinfo{volume}{33} (\bibinfo{year}{2020}), \bibinfo{pages}{9459--9474}.
\newblock


\bibitem[Li et~al\mbox{.}(2023)]%
        {li2023towards}
\bibfield{author}{\bibinfo{person}{Haitao Li}, \bibinfo{person}{Jia Chen}, \bibinfo{person}{Weihang Su}, \bibinfo{person}{Qingyao Ai}, {and} \bibinfo{person}{Yiqun Liu}.} \bibinfo{year}{2023}\natexlab{}.
\newblock \showarticletitle{Towards better web search performance: pre-training, fine-tuning and learning to rank}.
\newblock \bibinfo{journal}{\emph{arXiv preprint arXiv:2303.04710}} (\bibinfo{year}{2023}).
\newblock


\bibitem[Liu et~al\mbox{.}(2024a)]%
        {liu2024deepseek}
\bibfield{author}{\bibinfo{person}{Aixin Liu}, \bibinfo{person}{Bei Feng}, \bibinfo{person}{Bing Xue}, \bibinfo{person}{Bingxuan Wang}, \bibinfo{person}{Bochao Wu}, \bibinfo{person}{Chengda Lu}, \bibinfo{person}{Chenggang Zhao}, \bibinfo{person}{Chengqi Deng}, \bibinfo{person}{Chenyu Zhang}, \bibinfo{person}{Chong Ruan}, {et~al\mbox{.}}} \bibinfo{year}{2024}\natexlab{a}.
\newblock \showarticletitle{Deepseek-v3 technical report}.
\newblock \bibinfo{journal}{\emph{arXiv preprint arXiv:2412.19437}} (\bibinfo{year}{2024}).
\newblock


\bibitem[Liu et~al\mbox{.}(2024c)]%
        {liu2024ctrla}
\bibfield{author}{\bibinfo{person}{Huanshuo Liu}, \bibinfo{person}{Hao Zhang}, \bibinfo{person}{Zhijiang Guo}, \bibinfo{person}{Kuicai Dong}, \bibinfo{person}{Xiangyang Li}, \bibinfo{person}{Yi~Quan Lee}, \bibinfo{person}{Cong Zhang}, {and} \bibinfo{person}{Yong Liu}.} \bibinfo{year}{2024}\natexlab{c}.
\newblock \showarticletitle{CtrlA: Adaptive Retrieval-Augmented Generation via Probe-Guided Control}.
\newblock \bibinfo{journal}{\emph{arXiv preprint arXiv:2405.18727}} (\bibinfo{year}{2024}).
\newblock


\bibitem[Liu et~al\mbox{.}(2024b)]%
        {liu2024lost}
\bibfield{author}{\bibinfo{person}{Nelson~F Liu}, \bibinfo{person}{Kevin Lin}, \bibinfo{person}{John Hewitt}, \bibinfo{person}{Ashwin Paranjape}, \bibinfo{person}{Michele Bevilacqua}, \bibinfo{person}{Fabio Petroni}, {and} \bibinfo{person}{Percy Liang}.} \bibinfo{year}{2024}\natexlab{b}.
\newblock \showarticletitle{Lost in the middle: How language models use long contexts}.
\newblock \bibinfo{journal}{\emph{Transactions of the Association for Computational Linguistics}}  \bibinfo{volume}{12} (\bibinfo{year}{2024}), \bibinfo{pages}{157--173}.
\newblock


\bibitem[Ma et~al\mbox{.}(2023)]%
        {ma2023caseencoder}
\bibfield{author}{\bibinfo{person}{Yixiao Ma}, \bibinfo{person}{Yueyue Wu}, \bibinfo{person}{Weihang Su}, \bibinfo{person}{Qingyao Ai}, {and} \bibinfo{person}{Yiqun Liu}.} \bibinfo{year}{2023}\natexlab{}.
\newblock \showarticletitle{CaseEncoder: A Knowledge-enhanced Pre-trained Model for Legal Case Encoding}.
\newblock \bibinfo{journal}{\emph{arXiv preprint arXiv:2305.05393}} (\bibinfo{year}{2023}).
\newblock


\bibitem[Nanda et~al\mbox{.}(2023)]%
        {nanda2023fact}
\bibfield{author}{\bibinfo{person}{Neel Nanda}, \bibinfo{person}{Senthooran Rajamanoharan}, \bibinfo{person}{János Kramár}, {and} \bibinfo{person}{Rohin Shah}.} \bibinfo{year}{2023}\natexlab{}.
\newblock \bibinfo{booktitle}{\emph{Fact Finding: Attempting to Reverse-Engineer Factual Recall on the Neuron Level}}.
\newblock
\urldef\tempurl%
\url{https://www.lesswrong.com/posts/iGuwZTHWb6DFY3sKB/fact-finding-attempting-to-reverse-engineer-factual-recall}
\showURL{%
\tempurl}
\newblock
\shownote{Accessed: 2025-01-24}.


\bibitem[Rawte et~al\mbox{.}(2023)]%
        {rawte2023survey}
\bibfield{author}{\bibinfo{person}{Vipula Rawte}, \bibinfo{person}{Amit Sheth}, {and} \bibinfo{person}{Amitava Das}.} \bibinfo{year}{2023}\natexlab{}.
\newblock \showarticletitle{A survey of hallucination in large foundation models}.
\newblock \bibinfo{journal}{\emph{arXiv preprint arXiv:2309.05922}} (\bibinfo{year}{2023}).
\newblock


\bibitem[Robertson et~al\mbox{.}(2009)]%
        {robertson2009probabilistic}
\bibfield{author}{\bibinfo{person}{Stephen Robertson}, \bibinfo{person}{Hugo Zaragoza}, {et~al\mbox{.}}} \bibinfo{year}{2009}\natexlab{}.
\newblock \showarticletitle{The probabilistic relevance framework: BM25 and beyond}.
\newblock \bibinfo{journal}{\emph{Foundations and Trends{\textregistered} in Information Retrieval}} \bibinfo{volume}{3}, \bibinfo{number}{4} (\bibinfo{year}{2009}), \bibinfo{pages}{333--389}.
\newblock


\bibitem[Salemi and Zamani(2024)]%
        {salemi2024towards}
\bibfield{author}{\bibinfo{person}{Alireza Salemi} {and} \bibinfo{person}{Hamed Zamani}.} \bibinfo{year}{2024}\natexlab{}.
\newblock \showarticletitle{Towards a search engine for machines: Unified ranking for multiple retrieval-augmented large language models}. In \bibinfo{booktitle}{\emph{Proceedings of the 47th International ACM SIGIR Conference on Research and Development in Information Retrieval}}. \bibinfo{pages}{741--751}.
\newblock


\bibitem[Su(2025)]%
        {oneal2000homepage}
\bibfield{author}{\bibinfo{person}{Weihang Su}.} \bibinfo{year}{2025}\natexlab{}.
\newblock \bibinfo{title}{Homepage of Weihang Su}.
\newblock \bibinfo{howpublished}{\url{https://oneal2000.github.io/}}.
\newblock
\newblock
\shownote{Accessed: 2025-03-28}.


\bibitem[Su et~al\mbox{.}(2024a)]%
        {su2024wikiformer}
\bibfield{author}{\bibinfo{person}{Weihang Su}, \bibinfo{person}{Qingyao Ai}, \bibinfo{person}{Xiangsheng Li}, \bibinfo{person}{Jia Chen}, \bibinfo{person}{Yiqun Liu}, \bibinfo{person}{Xiaolong Wu}, {and} \bibinfo{person}{Shengluan Hou}.} \bibinfo{year}{2024}\natexlab{a}.
\newblock \showarticletitle{Wikiformer: Pre-training with structured information of wikipedia for ad-hoc retrieval}. In \bibinfo{booktitle}{\emph{Proceedings of the AAAI Conference on Artificial Intelligence}}, Vol.~\bibinfo{volume}{38}. \bibinfo{pages}{19026--19034}.
\newblock


\bibitem[Su et~al\mbox{.}(2023a)]%
        {su2023caseformer}
\bibfield{author}{\bibinfo{person}{Weihang Su}, \bibinfo{person}{Qingyao Ai}, \bibinfo{person}{Yueyue Wu}, \bibinfo{person}{Yixiao Ma}, \bibinfo{person}{Haitao Li}, {and} \bibinfo{person}{Yiqun Liu}.} \bibinfo{year}{2023}\natexlab{a}.
\newblock \showarticletitle{Caseformer: Pre-training for Legal Case Retrieval}.
\newblock \bibinfo{journal}{\emph{arXiv preprint arXiv:2311.00333}} (\bibinfo{year}{2023}).
\newblock


\bibitem[Su et~al\mbox{.}(2024b)]%
        {su2024stard}
\bibfield{author}{\bibinfo{person}{Weihang Su}, \bibinfo{person}{Yiran Hu}, \bibinfo{person}{Anzhe Xie}, \bibinfo{person}{Qingyao Ai}, \bibinfo{person}{Quezi Bing}, \bibinfo{person}{Ning Zheng}, \bibinfo{person}{Yun Liu}, \bibinfo{person}{Weixing Shen}, {and} \bibinfo{person}{Yiqun Liu}.} \bibinfo{year}{2024}\natexlab{b}.
\newblock \showarticletitle{{STARD}: A {C}hinese Statute Retrieval Dataset Derived from Real-life Queries by Non-professionals}. In \bibinfo{booktitle}{\emph{Findings of the Association for Computational Linguistics: EMNLP 2024}}, \bibfield{editor}{\bibinfo{person}{Yaser Al-Onaizan}, \bibinfo{person}{Mohit Bansal}, {and} \bibinfo{person}{Yun-Nung Chen}} (Eds.). \bibinfo{publisher}{Association for Computational Linguistics}, \bibinfo{address}{Miami, Florida, USA}, \bibinfo{pages}{10658--10671}.
\newblock
\urldef\tempurl%
\url{https://doi.org/10.18653/v1/2024.findings-emnlp.625}
\showDOI{\tempurl}


\bibitem[Su et~al\mbox{.}(2023b)]%
        {su2023thuir2}
\bibfield{author}{\bibinfo{person}{Weihang Su}, \bibinfo{person}{Xiangsheng Li}, \bibinfo{person}{Yiqun Liu}, \bibinfo{person}{Min Zhang}, {and} \bibinfo{person}{Shaoping Ma}.} \bibinfo{year}{2023}\natexlab{b}.
\newblock \showarticletitle{Thuir2 at ntcir-16 session search (ss) task}.
\newblock \bibinfo{journal}{\emph{arXiv preprint arXiv:2307.00250}} (\bibinfo{year}{2023}).
\newblock


\bibitem[Su et~al\mbox{.}(2024c)]%
        {su2024mitigating}
\bibfield{author}{\bibinfo{person}{Weihang Su}, \bibinfo{person}{Yichen Tang}, \bibinfo{person}{Qingyao Ai}, \bibinfo{person}{Changyue Wang}, \bibinfo{person}{Zhijing Wu}, {and} \bibinfo{person}{Yiqun Liu}.} \bibinfo{year}{2024}\natexlab{c}.
\newblock \showarticletitle{Mitigating entity-level hallucination in large language models}. In \bibinfo{booktitle}{\emph{Proceedings of the 2024 Annual International ACM SIGIR Conference on Research and Development in Information Retrieval in the Asia Pacific Region}}. \bibinfo{pages}{23--31}.
\newblock


\bibitem[Su et~al\mbox{.}(2024d)]%
        {su2024dragin}
\bibfield{author}{\bibinfo{person}{Weihang Su}, \bibinfo{person}{Yichen Tang}, \bibinfo{person}{Qingyao Ai}, \bibinfo{person}{Zhijing Wu}, {and} \bibinfo{person}{Yiqun Liu}.} \bibinfo{year}{2024}\natexlab{d}.
\newblock \showarticletitle{{DRAGIN}: Dynamic Retrieval Augmented Generation based on the Real-time Information Needs of Large Language Models}. In \bibinfo{booktitle}{\emph{Proceedings of the 62nd Annual Meeting of the Association for Computational Linguistics (Volume 1: Long Papers)}}, \bibfield{editor}{\bibinfo{person}{Lun-Wei Ku}, \bibinfo{person}{Andre Martins}, {and} \bibinfo{person}{Vivek Srikumar}} (Eds.). \bibinfo{publisher}{Association for Computational Linguistics}, \bibinfo{address}{Bangkok, Thailand}, \bibinfo{pages}{12991--13013}.
\newblock
\urldef\tempurl%
\url{https://doi.org/10.18653/v1/2024.acl-long.702}
\showDOI{\tempurl}


\bibitem[Su et~al\mbox{.}(2025a)]%
        {su2025paramrag}
\bibfield{author}{\bibinfo{person}{Weihang Su}, \bibinfo{person}{Yichen Tang}, \bibinfo{person}{Qingyao Ai}, \bibinfo{person}{Junxi Yan}, \bibinfo{person}{Changyue Wang}, \bibinfo{person}{Hongning Wang}, \bibinfo{person}{Ziyi Ye}, \bibinfo{person}{Yujia Zhou}, {and} \bibinfo{person}{Yiqun Liu}.} \bibinfo{year}{2025}\natexlab{a}.
\newblock \showarticletitle{Parametric Retrieval-Augmented Generation}.
\newblock \bibinfo{journal}{\emph{arXiv preprint arXiv:2501.15915}} (\bibinfo{year}{2025}).
\newblock


\bibitem[Su et~al\mbox{.}(2024e)]%
        {su2024unsupervised}
\bibfield{author}{\bibinfo{person}{Weihang Su}, \bibinfo{person}{Changyue Wang}, \bibinfo{person}{Qingyao Ai}, \bibinfo{person}{Yiran Hu}, \bibinfo{person}{Zhijing Wu}, \bibinfo{person}{Yujia Zhou}, {and} \bibinfo{person}{Yiqun Liu}.} \bibinfo{year}{2024}\natexlab{e}.
\newblock \showarticletitle{Unsupervised Real-Time Hallucination Detection based on the Internal States of Large Language Models}. In \bibinfo{booktitle}{\emph{Findings of the Association for Computational Linguistics: ACL 2024}}, \bibfield{editor}{\bibinfo{person}{Lun-Wei Ku}, \bibinfo{person}{Andre Martins}, {and} \bibinfo{person}{Vivek Srikumar}} (Eds.). \bibinfo{publisher}{Association for Computational Linguistics}, \bibinfo{address}{Bangkok, Thailand}, \bibinfo{pages}{14379--14391}.
\newblock
\urldef\tempurl%
\url{https://doi.org/10.18653/v1/2024.findings-acl.854}
\showDOI{\tempurl}


\bibitem[Su et~al\mbox{.}(2024f)]%
        {su2023legalaid}
\bibfield{author}{\bibinfo{person}{Weihang Su}, \bibinfo{person}{Changyue Wang}, \bibinfo{person}{Anzhe Xie}, \bibinfo{person}{Qingyao Ai}, \bibinfo{person}{Yiran Hu}, {and} \bibinfo{person}{Yiqun Liu}.} \bibinfo{year}{2024}\natexlab{f}.
\newblock \bibinfo{title}{LegalAID: A Large Language Model for the Chinese Legal Field}.
\newblock \bibinfo{howpublished}{\url{https://github.com/oneal2000/LegalAID}}.
\newblock


\bibitem[Su et~al\mbox{.}(2025b)]%
        {su2025judge}
\bibfield{author}{\bibinfo{person}{Weihang Su}, \bibinfo{person}{Baoqing Yue}, \bibinfo{person}{Qingyao Ai}, \bibinfo{person}{Yiran Hu}, \bibinfo{person}{Jiaqi Li}, \bibinfo{person}{Changyue Wang}, \bibinfo{person}{Kaiyuan Zhang}, \bibinfo{person}{Yueyue Wu}, {and} \bibinfo{person}{Yiqun Liu}.} \bibinfo{year}{2025}\natexlab{b}.
\newblock \showarticletitle{JuDGE: Benchmarking Judgment Document Generation for Chinese Legal System}. In \bibinfo{booktitle}{\emph{Proceedings of the 48th International ACM SIGIR Conference on Research and Development in Information Retrieval (SIGIR '25), July 13--18, 2025, Padua, Italy}}.
\newblock
\urldef\tempurl%
\url{https://doi.org/10.1145/3726302.3730295}
\showDOI{\tempurl}


\bibitem[Tan et~al\mbox{.}(2025)]%
        {tan2025better}
\bibfield{author}{\bibinfo{person}{Yuqiao Tan}, \bibinfo{person}{Shizhu He}, \bibinfo{person}{Huanxuan Liao}, \bibinfo{person}{Jun Zhao}, {and} \bibinfo{person}{Kang Liu}.} \bibinfo{year}{2025}\natexlab{}.
\newblock \showarticletitle{Better wit than wealth: Dynamic Parametric Retrieval Augmented Generation for Test-time Knowledge Enhancement}.
\newblock \bibinfo{journal}{\emph{arXiv preprint arXiv:2503.23895}} (\bibinfo{year}{2025}).
\newblock


\bibitem[Touvron et~al\mbox{.}(2023)]%
        {touvron2023llama}
\bibfield{author}{\bibinfo{person}{Hugo Touvron}, \bibinfo{person}{Thibaut Lavril}, \bibinfo{person}{Gautier Izacard}, \bibinfo{person}{Xavier Martinet}, \bibinfo{person}{Marie-Anne Lachaux}, \bibinfo{person}{Timoth{\'e}e Lacroix}, \bibinfo{person}{Baptiste Rozi{\`e}re}, \bibinfo{person}{Naman Goyal}, \bibinfo{person}{Eric Hambro}, \bibinfo{person}{Faisal Azhar}, {et~al\mbox{.}}} \bibinfo{year}{2023}\natexlab{}.
\newblock \showarticletitle{Llama: Open and efficient foundation language models}.
\newblock \bibinfo{journal}{\emph{arXiv preprint arXiv:2302.13971}} (\bibinfo{year}{2023}).
\newblock


\bibitem[Tu et~al\mbox{.}(2025)]%
        {tu2025rbft}
\bibfield{author}{\bibinfo{person}{Yiteng Tu}, \bibinfo{person}{Weihang Su}, \bibinfo{person}{Yujia Zhou}, \bibinfo{person}{Yiqun Liu}, {and} \bibinfo{person}{Qingyao Ai}.} \bibinfo{year}{2025}\natexlab{}.
\newblock \showarticletitle{RbFT: Robust Fine-tuning for Retrieval-Augmented Generation against Retrieval Defects}.
\newblock \bibinfo{journal}{\emph{arXiv preprint arXiv:2501.18365}} (\bibinfo{year}{2025}).
\newblock


\bibitem[Wang et~al\mbox{.}(2024a)]%
        {wang2024knowledge}
\bibfield{author}{\bibinfo{person}{Changyue Wang}, \bibinfo{person}{Weihang Su}, \bibinfo{person}{Qingyao Ai}, {and} \bibinfo{person}{Yiqun Liu}.} \bibinfo{year}{2024}\natexlab{a}.
\newblock \showarticletitle{Knowledge Editing through Chain-of-Thought}.
\newblock \bibinfo{journal}{\emph{arXiv preprint arXiv:2412.17727}} (\bibinfo{year}{2024}).
\newblock


\bibitem[Wang et~al\mbox{.}(2025a)]%
        {wang2025decouple}
\bibfield{author}{\bibinfo{person}{Changyue Wang}, \bibinfo{person}{Weihang Su}, \bibinfo{person}{Qingyao Ai}, {and} \bibinfo{person}{Yiqun Liu}.} \bibinfo{year}{2025}\natexlab{a}.
\newblock \showarticletitle{Decoupling Reasoning and Knowledge Injection for In-Context Knowledge Editing}.
\newblock \bibinfo{journal}{\emph{arXiv preprint arXiv:2506.00536}} (\bibinfo{year}{2025}).
\newblock


\bibitem[Wang et~al\mbox{.}(2025b)]%
        {wang2025joint}
\bibfield{author}{\bibinfo{person}{Changyue Wang}, \bibinfo{person}{Weihang Su}, \bibinfo{person}{Qingyao Ai}, {and} \bibinfo{person}{Yiqun Liu}.} \bibinfo{year}{2025}\natexlab{b}.
\newblock \showarticletitle{Joint Evaluation of Answer and Reasoning Consistency for Hallucination Detection in Large Reasoning Models}.
\newblock \bibinfo{journal}{\emph{arXiv preprint arXiv:2506.04832}} (\bibinfo{year}{2025}).
\newblock


\bibitem[Wang et~al\mbox{.}(2024b)]%
        {wang2024lekube}
\bibfield{author}{\bibinfo{person}{Changyue Wang}, \bibinfo{person}{Weihang Su}, \bibinfo{person}{Hu Yiran}, \bibinfo{person}{Qingyao Ai}, \bibinfo{person}{Yueyue Wu}, \bibinfo{person}{Cheng Luo}, \bibinfo{person}{Yiqun Liu}, \bibinfo{person}{Min Zhang}, {and} \bibinfo{person}{Shaoping Ma}.} \bibinfo{year}{2024}\natexlab{b}.
\newblock \showarticletitle{LeKUBE: A Legal Knowledge Update BEnchmark}.
\newblock \bibinfo{journal}{\emph{arXiv preprint arXiv:2407.14192}} (\bibinfo{year}{2024}).
\newblock


\bibitem[Wang et~al\mbox{.}(2024c)]%
        {wang2024knowledgesurvey}
\bibfield{author}{\bibinfo{person}{Song Wang}, \bibinfo{person}{Yaochen Zhu}, \bibinfo{person}{Haochen Liu}, \bibinfo{person}{Zaiyi Zheng}, \bibinfo{person}{Chen Chen}, {and} \bibinfo{person}{Jundong Li}.} \bibinfo{year}{2024}\natexlab{c}.
\newblock \showarticletitle{Knowledge editing for large language models: A survey}.
\newblock \bibinfo{journal}{\emph{Comput. Surveys}} \bibinfo{volume}{57}, \bibinfo{number}{3} (\bibinfo{year}{2024}), \bibinfo{pages}{1--37}.
\newblock


\bibitem[Yang et~al\mbox{.}(2024)]%
        {yang2024qwen2}
\bibfield{author}{\bibinfo{person}{An Yang}, \bibinfo{person}{Baosong Yang}, \bibinfo{person}{Beichen Zhang}, \bibinfo{person}{Binyuan Hui}, \bibinfo{person}{Bo Zheng}, \bibinfo{person}{Bowen Yu}, \bibinfo{person}{Chengyuan Li}, \bibinfo{person}{Dayiheng Liu}, \bibinfo{person}{Fei Huang}, \bibinfo{person}{Haoran Wei}, {et~al\mbox{.}}} \bibinfo{year}{2024}\natexlab{}.
\newblock \showarticletitle{Qwen2. 5 Technical Report}.
\newblock \bibinfo{journal}{\emph{arXiv preprint arXiv:2412.15115}} (\bibinfo{year}{2024}).
\newblock


\bibitem[Yao et~al\mbox{.}(2024)]%
        {yao2024seakr}
\bibfield{author}{\bibinfo{person}{Zijun Yao}, \bibinfo{person}{Weijian Qi}, \bibinfo{person}{Liangming Pan}, \bibinfo{person}{Shulin Cao}, \bibinfo{person}{Linmei Hu}, \bibinfo{person}{Weichuan Liu}, \bibinfo{person}{Lei Hou}, {and} \bibinfo{person}{Juanzi Li}.} \bibinfo{year}{2024}\natexlab{}.
\newblock \showarticletitle{Seakr: Self-aware knowledge retrieval for adaptive retrieval augmented generation}.
\newblock \bibinfo{journal}{\emph{arXiv preprint arXiv:2406.19215}} (\bibinfo{year}{2024}).
\newblock


\bibitem[Ye et~al\mbox{.}(2024)]%
        {ye2024relevance}
\bibfield{author}{\bibinfo{person}{Ziyi Ye}, \bibinfo{person}{Xiaohui Xie}, \bibinfo{person}{Qingyao Ai}, \bibinfo{person}{Yiqun Liu}, \bibinfo{person}{Zhihong Wang}, \bibinfo{person}{Weihang Su}, {and} \bibinfo{person}{Min Zhang}.} \bibinfo{year}{2024}\natexlab{}.
\newblock \showarticletitle{Relevance Feedback with Brain Signals}.
\newblock \bibinfo{journal}{\emph{ACM Transactions on Information Systems}} \bibinfo{volume}{42}, \bibinfo{number}{4} (\bibinfo{year}{2024}), \bibinfo{pages}{1--37}.
\newblock


\bibitem[Yu and Ananiadou(2024)]%
        {yu-ananiadou-2024-neuron}
\bibfield{author}{\bibinfo{person}{Zeping Yu} {and} \bibinfo{person}{Sophia Ananiadou}.} \bibinfo{year}{2024}\natexlab{}.
\newblock \showarticletitle{Neuron-Level Knowledge Attribution in Large Language Models}. In \bibinfo{booktitle}{\emph{Proceedings of the 2024 Conference on Empirical Methods in Natural Language Processing}}, \bibfield{editor}{\bibinfo{person}{Yaser Al-Onaizan}, \bibinfo{person}{Mohit Bansal}, {and} \bibinfo{person}{Yun-Nung Chen}} (Eds.). \bibinfo{publisher}{Association for Computational Linguistics}, \bibinfo{address}{Miami, Florida, USA}, \bibinfo{pages}{3267--3280}.
\newblock
\urldef\tempurl%
\url{https://doi.org/10.18653/v1/2024.emnlp-main.191}
\showDOI{\tempurl}


\bibitem[Zhai(2008)]%
        {zhai2008statistical}
\bibfield{author}{\bibinfo{person}{ChengXiang Zhai}.} \bibinfo{year}{2008}\natexlab{}.
\newblock \showarticletitle{Statistical language models for information retrieval}.
\newblock \bibinfo{journal}{\emph{Synthesis lectures on human language technologies}} \bibinfo{volume}{1}, \bibinfo{number}{1} (\bibinfo{year}{2008}), \bibinfo{pages}{1--141}.
\newblock


\bibitem[Zhan(2025)]%
        {jingtaozhanhomepage}
\bibfield{author}{\bibinfo{person}{Jingtao Zhan}.} \bibinfo{year}{2025}\natexlab{}.
\newblock \bibinfo{title}{Homepage of Jingtao Zhan}.
\newblock \bibinfo{howpublished}{\url{https://jingtaozhan.github.io/}}.
\newblock
\newblock
\shownote{Accessed: 2025-03-28}.


\bibitem[Zhan et~al\mbox{.}(2021)]%
        {zhan2021optimizing}
\bibfield{author}{\bibinfo{person}{Jingtao Zhan}, \bibinfo{person}{Jiaxin Mao}, \bibinfo{person}{Yiqun Liu}, \bibinfo{person}{Jiafeng Guo}, \bibinfo{person}{Min Zhang}, {and} \bibinfo{person}{Shaoping Ma}.} \bibinfo{year}{2021}\natexlab{}.
\newblock \showarticletitle{Optimizing dense retrieval model training with hard negatives}. In \bibinfo{booktitle}{\emph{Proceedings of the 44th international ACM SIGIR conference on research and development in information retrieval}}. \bibinfo{pages}{1503--1512}.
\newblock


\end{thebibliography}

\end{document}